\documentclass[letterpaper, 10 pt, conference]{ieeeconf}  % Comment this line out if you need a4paper

\IEEEoverridecommandlockouts                              % This command is only needed if 
\usepackage{graphicx} % Required for inserting images
\usepackage{xcolor}
\usepackage{amsmath}
\usepackage{amssymb}
\usepackage[font=small]{caption}
\usepackage{cite}
\usepackage{booktabs}
\usepackage{tabulary}
\usepackage{stfloats}
\usepackage{makecell}
\usepackage{subcaption}
\usepackage{hyperref}
\usepackage{wrapfig}

\title{ROSflight 2.0: Lean ROS 2-Based Autopilot for Unmanned Aerial Vehicles}
\author{Jacob Moore$^1$, Phil Tokumaru$^2$, Ian Reid$^1$, Brandon Sutherland$^1$, Joseph Ritchie$^1$, Gabe Snow$^1$, Tim McLain$^1$% <-this % stops a space
\thanks{$^{1}$Brigham Young University}
\thanks{$^{2}$AeroVironment Inc.}
}

\begin{document}

\maketitle

\begin{abstract}
ROSflight is a lean, open-source autopilot ecosystem for unmanned aerial vehicles (UAVs).
Designed by researchers for researchers, it is built to lower the barrier to entry to UAV research and accelerate the transition from simulation to hardware experiments by maintaining a lean (not full-featured), well-documented, and modular codebase.
This publication builds on previous treatments and describes significant additions to the architecture that improve the modularity and usability of ROSflight, including the transition from ROS 1 to ROS 2, supported hardware, low-level actuator mixing, and the simulation environment.
We believe that these changes improve the usability of ROSflight and enable ROSflight to accelerate research in areas like advanced-air mobility.
Hardware results are provided, showing that ROSflight is able to control a multirotor over a serial connection at 400 Hz while closing all control loops on the companion computer.
\end{abstract}

\section{Introduction} \label{Introduction}
In recent years, interest in unmanned aerial vehicles (UAVs) has increased significantly.
Technological advances have enabled numerous applications of UAVs, including package delivery, photography, search-and-rescue, firefighting, as well as military applications.

Advanced air mobility (AAM), a category broadly referring to increasing autonomy in urban areas for civilian use, is also currently an area of high interest.
AAM aircraft often take the form of eVTOL aircraft, and specialized autopilots, algorithms, and hardware are needed to effectively conduct research in this field.
Because of this, researchers often need access to the inner workings of an autopilot (e.g., the state estimator or inner loop controller), which makes commercial closed-source autopilots or many open-source autopilots difficult to use.
Additionally, simulation and hardware experiments are critical in AAM research to ensure proposed systems and methodologies are safe and function as intended.

ROSflight is a lean, open-source autopilot for UAVs built for research.
Because it is designed to be lean, ROSflight is not full-featured and does not boast many of the state-of-the-art functions available in other popular open-source autopilots \cite{px4,ardupilot}.
Instead, ROSflight offers only the basic functionality needed to support UAV research, prioritizing understandability.
While this places more responsibility on the end user to develop application-specific code, we believe a lean architecture reduces the \textit{black-box} nature of the autopilot, thus reducing the total effort to implement a user's application code.
Additionally, the ROSflight project has a strong emphasis on clear code and complete documentation, which improves accessibility and lowers the barrier to entry for researchers and students alike.
Documentation can be found on the project website, \href{rosflight.org}{rosflight.org}.

Built on the Robot Operating System (ROS 2) \cite{ros2_2022}, ROSflight is designed with a modular architecture to fit the needs of varying airframes and applications.
ROSflight is designed to enable true software in the loop (SIL) simulation.
When using ROSflight, the same code that runs the autopilot in simulation also flies the vehicle in hardware, with no changes---significantly enhancing the transition from simulation to hardware experiments.

ROSflight has received detailed treatment in \cite{rosflight2016,rosflight2020}.
We rely on \cite{rosflight2020} for an excellent description of the overall architecture and design goals of the ROSflight project.
The contributions of this work are to describe significant improvements to ROSflight in the release of ROSflight 2.0 including 
\begin{itemize}
    \item the transition from ROS 1 to ROS 2,
    \item improved modularity and accuracy in the actuator mixing,
    \item new supported hardware for faster and more reliable operation,
    \item a restructured and modular simulation environment to support diverse simulation needs, and
    \item flight test results demonstrating these improvements in hardware.
\end{itemize}

Due to these advancements, ROSflight 2.0 improves the modularity and usability of the software, enabling ROSflight to accelerate research in areas like advanced air mobility.

The rest of the paper is organized as follows:
Section \ref{related-work} describes work similar to ROSflight.
A brief overview of ROSflight is described in Section \ref{sec:overview}.
Improvements to the ROSflight architecture are described in detail in Section \ref{software-updates}, and supported hardware and improvements to the simulation environment are discussed in Section \ref{hardware-updates}.
Hardware results demonstrating these improvements to ROSflight are described in Section \ref{hardware-demos} and we conclude in Section \ref{summary}.

\section{Related Work}\label{related-work}
Advances in UAV technology have given rise to many excellent and mature open-source autopilots like PX4 \cite{px4} and ArduPilot \cite{ardupilot}.
Both of these autopilots offer impressive feature sets, state-of-the-art performance, extensive community support and adoption, and \textit{plug-and-play} functionality.
Additionally, both projects have good support for ROS 2, thus improving integration of external code libraries into an autopilot \cite{open-source-autopilot-survey}.
However, the large code bases, complexity, and steep learning curve of these autopilots decrease understandability.
This can increase the time and effort required to integrate external code and conduct simulation and hardware tests. 
As noted by D'Angelo et al., these challenges often cause researchers to prefer developing custom firmware from scratch \cite{DAngelo2024_ICUAS}.

ROSflight is primarily designed to be understandable with clean, lean code and complete documentation.
Since ROSflight is lean, it does not offer most of the features offered by \cite{px4,ardupilot}, and therefore is not suitable for commercial applications or production-ready vehicles.
Instead, ROSflight targets research, classroom, or early-stage development projects where understandability and ease of modification is crucial.

Additionally, ROSflight moves most of the autopilot stack to a Linux-based companion computer which has more compute power and is easier to develop and rapidly test than an embedded microcontroller.
ROSflight also emphasizes true software-in-the-loop simulation \cite{rosflight2020}, meaning that the core software that runs in simulation also runs in hardware, with no changes.
Other open-source autopilots \cite{paparazzi,inav} are less mature, have limited flexibility and ROS 2 support, or are designed primarily for hobbyist use.
ROSflight is built around ROS 2 and is explicitly designed to fill researchers' needs.

\section{ROSflight Overview}\label{sec:overview}
\begin{figure}[tbp]
    \centering
    \includegraphics[width=0.75\linewidth]{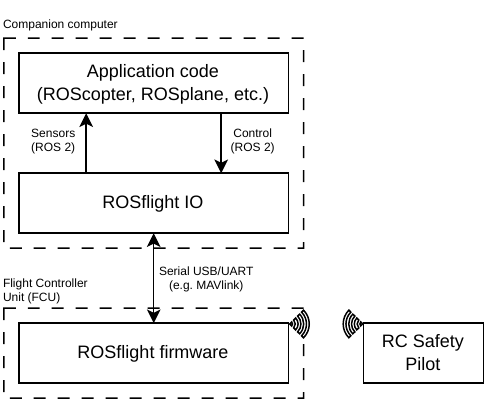}
    \caption{Overview of the ROSflight architecture.}
    \label{fig:rf-overview}
\end{figure}

An overview of the ROSflight architecture is shown in Figure \ref{fig:rf-overview}.
At a high level, it is comprised of two onboard systems: a low-level embedded microcontroller (called the flight control unit) and a Linux-based companion computer (also called the flight computer).

\noindent
\textbf{Flight control unit (FCU):}
The FCU interfaces directly with the physical aircraft and communicates with a Linux companion computer using a high-speed UART or USB CDC ACM serial interface.
This FCU runs the ROSflight firmware, and is responsible for collecting sensor measurements (a \textit{sensor aggregator}), controlling actuators, streaming information to the companion computer, and monitoring the state of the software (e.g. armed or disarmed). 
The FCU also receives radio control (RC) commands directly from a safety pilot via an RC receiver.
An estimator and controller are present on the firmware to manage the fast inner loops of the control architecture (e.g., angle and rate loops for a multirotor).
Note that while the vehicle is fully controllable by the safety pilot without the companion computer, the majority of the autonomy stack is found on the companion computer, so only limited autonomy is available without it.

\noindent
\textbf{Companion computer:}
The companion computer, running Linux, receives sensor measurements and sends actuator commands to the FCU over the serial interface.
The serial protocol is abstracted in such a way that different serial protocols can be used, and MAVlink is used as the default serial communication protocol.
Most of the autonomy stack, including high-level state estimation and control algorithms, resides on this computer.
All modules on the companion computer are written as ROS 2 nodes, enhancing the modularity of the system.
When connected on the same network using standard WiFi or another network, the companion computer can communicate with other computers like a ground-station laptop so users can easily monitor the system.

The subject of this work is to describe the significant improvements to both the hardware and software associated with the FCU.
Examples of out-of-the-box high-level autonomy stacks using ROSflight are available with ROSplane\footnote{https://github.com/rosflight/rosplane} \cite{rosplane2017} and ROScopter\footnote{https://github.com/rosflight/roscopter}.

\section{Software Design}\label{software-updates}
\subsection{ROS 1 to ROS 2}
Previous versions of ROSflight used ROS 1, which is no longer supported by the maintainers of ROS.
ROSflight now uses ROS 2, and primarily supports long-term support (LTS) versions of ROS 2, which are ROS 2 Humble with Ubuntu 22.04 and ROS 2 Jazzy with Ubuntu 24.04.
ROS 2 offers several improvements over the previous ROS 1 system, making it more robust and reliable \cite{ros2_2022}.
The core ROSflight firmware on the embedded microcontroller does not use ROS (e.g. micro-ROS \cite{microROS_2023}), and instead communicates with the ROS 2-based I/O node (called ROSflightIO) on the companion computer over a serial connection.
The ROSflight simulation is built around ROS 2, and exploits this to improve modularity as will be discussed in Section \ref{sec:simulation}.

\subsection{Mixer}\label{mixer-section}
\begin{figure}
    \centering
    \includegraphics[width=0.8\columnwidth]{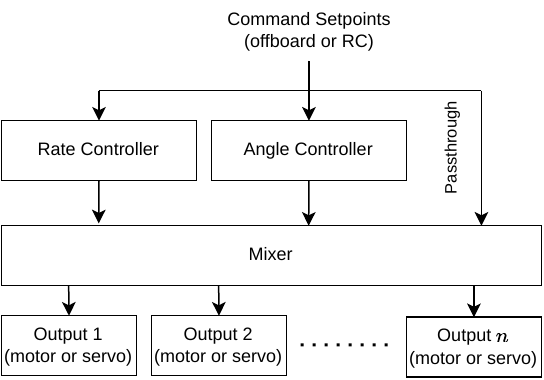}
    \caption{Diagram of how the ROSflight mixer interfaces with other ROSflight modules. The mixer takes in either the output of a controller or setpoints that bypass the controller, called \textit{pass-through} commands, and outputs actuator signals. The setpoints are created by the onboard computer or the RC safety pilot.}
    \label{fig:mixer-diagram}
\end{figure}

In ROSflight 2.0, the mixer has been restructured to offer greater flexibility and control.
The mixer is a component in the software responsible for taking in a controller command and transforming it to actuator outputs.
This process is called \textit{actuator mixing} or \textit{force allocation}.
Given a desired command, the mixer allocates actuator effort to achieve it as closely as possible.
The inputs to the mixer are the output of a controller, as shown in the block diagram in Figure \ref{fig:mixer-diagram}.
Similar to \cite{uavbook}, we formulate a linear mapping between the inputs and outputs using a mixing matrix, which takes the form in Equation (\ref{eq:mixer}).
\begin{equation}\label{eq:mixer}
    \tau = M^\dagger u,
\end{equation}
where $\tau \in \mathbb{R}^n$ is a vector of commands that is eventually written to the actuators (after any necessary scaling and saturation operations), $M \in \mathbb{R}^{m \times n}$ is the mixing matrix, $u \in \mathbb{R}^m$ is the vector of controller commands, and $(\cdot)^\dagger$ is the Moore-Penrose pseudoinverse.
In ROSflight, $u$ is composed of either RC safety pilot commands or \textit{offboard commands} from the companion computer (offboard with respect to the FCU), as described in \cite{rosflight2020}.

In ROSflight, we let $m=10$ and $n=10$, meaning there are ten output channels corresponding to hardware outputs (PWM or DShot) on the FCU, and ten desired commands to the mixer, requiring 100 values.
Additionally, the ROSflight mixer defines a PWM rate and an output type (motor, servo, GPIO, or auxiliary) for each output channel, yielding an additional 20 values.
Note that the actual number of PWM output pins is limited by the FCU hardware.
If a hardware board has more than ten PWM output pins, the remainder are automatically defined to be auxiliary outputs, which are simply PWM outputs that do not pass through the mixer.

The authors in \cite{uavbook} define $u$ to be a vector of the desired forces, $F$, and torques, $Q$, that are to be applied to the physical system.
However, the semantic meaning of the mixer input vector $u$ is closely tied with the elements of the mixer matrix.
Thus, ROSflight does not constrain the input vector $u$ to be a vector of forces and torques, but instead a vector of ten generic control inputs, all of which may not be used in a given setting.
For example, a quadrotor mixer may take in desired forces and torques ($F_z, Q_x, Q_y,Q_z$), corresponding to four of the six degrees of freedom of the physical aircraft, and return individual motor commands, while a V-tail fixed-wing mixer may take in desired control surface commands ($\delta_a, \delta_e, \delta_t, \delta_r$) and return individual servo commands.
An example of Equation (\ref{eq:mixer}) for a V-tail fixed-wing mixer is

\begin{equation*}\label{eq:example_mixer}
    \begin{bmatrix}
        \delta_a \\ \delta_{lr} \\ \delta_{rr} \\ \delta_t \\ 0_{6\times1}
    \end{bmatrix}
    =
    \begin{bmatrix}
        1 &    0 &   0 & 0 & 0_{1\times6}\\
        0 & -0.5 & 0.5 & 0 & 0_{1\times6}\\
        0 &  0.5 & 0.5 & 0 & 0_{1\times6}\\
        0 &    0 &   0 & 1 & 0_{1\times6}\\
        0_{6\times1} & 0_{6\times1} & 0_{6\times1} & 0_{6\times1} & 0_{6\times6}
    \end{bmatrix}
    \begin{bmatrix}
        \delta_a^c \\ \delta_e^c \\ \delta_r^c \\ \delta_t^c \\ 0_{6\times1}
    \end{bmatrix},
\end{equation*}
where $u = [\delta_a^c, \delta_e^c, \delta_r^c, \delta_t^c, 0_{1\times 6}]^T$ is the input vector containing the desired aileron, elevator, rudder, and throttle commands, respectively, and $\tau = [\delta_a, \delta_{lr}, \delta_{rr}, \delta_t, 0_{1\times 6}]^T$ is the output vector containing the commands to the servos controlling the ailerons, left ruddervator, right ruddervator, and throttle, respectively.
Since in this example the remaining six control inputs in $u$ are not used, they are zero.
Note that the signs of the coefficients in the mixing matrix are dependent on the orientation of the control-surface servos on the aircraft.

The command vector $u$ is not constrained to be an ordered vector of forces and torques, so ROSflight must determine how to map RC inputs to the $u$ command vector.
To do this, ROSflight assumes that the RC command vector takes the form $u=[F_x,F_y,F_z,Q_x,Q_y,Q_z, 0_{1\times4}]^T$, where the $F$ channels correspond to force or throttle RC inputs.
The $Q$ channels correspond to torque or rotational RC inputs (e.g. roll torque, angle, or rate commands).

Since the values of the elements of a mixing matrix are highly dependent on the physical airframe and the output of the controller, the ROSflight mixer has been re-designed to offer additional flexibility.
Key features allow users to
\begin{itemize}
    \item use predefined mixers for commonly used aircraft,
    \item define a custom mixer loaded at runtime,
    \item specify a separate mixer for the RC safety pilot and the onboard computer, and
    \item post-process the output vector $\tau$ with empirical motor parameters for a higher-fidelity model.
\end{itemize}

\subsubsection{Predefined Mixers}
Some predefined, hard-coded mixers are included for convenience in the implementation of the ROSflight firmware.
These mixers can be selected at runtime via parameters, and are available for standard airframes, including quadrotor, hexarotor, standard fixed-wing, or V-tail fixed-wing airframes.
A full list of predefined mixers is available on \href{rosflight.org}{rosflight.org}.
Each predefined mixer definition defines the type of each output channel (e.g., motor, servo, GPIO, or auxiliary), the PWM rate assigned to each channel, and the values of either $M$ or $M^\dagger$.
See \cite{rosflight_arxiv} for the derivation and assumptions of the predefined mixers.

\subsubsection{Custom Mixer}
Since the mixer is intimately tied to the physical geometry of the aircraft, users may need mixers not included in the predefined mixers hard-coded in ROSflight.
Thus, ROSflight includes the ability for users to load their own custom mixing matrix values via parameters.
A custom mixer must include both the 100 mixing matrix parameters as well as the 20 header parameters.
The names of these parameters and default values can be found on \href{rosflight.org}{rosflight.org}.

A custom mixer grants additional flexibility to ROSflight, allowing users to employ a higher fidelity model than the predefined mixers or to design mixers for nonstandard aircraft (e.g., eVTOL) without having to reflash the firmware.

\subsubsection{Pass-through Mixer}
ROSflight can be operated in \textit{pass-through} mode, where offboard commands bypass the firmware controller and progress directly to the mixer, as shown in Figure (\ref{fig:mixer-diagram}).
The high-speed serial connection enables sending these pass-through offboard commands at a high rate, thus giving direct access to individual motors at a high rate.
Such a configuration enables low-level control of a vehicle from the companion computer. 
Control over this low-level behavior is advantageous when developing novel flight controllers; however, this is not always possible with state-of-the-art autopilots like PX4 without significant changes to the firmware and navigating complex dependencies \cite{hegre2025neuralnetworkmodepx4}.
Additionally, PX4 explicitly disallows pass-through control in some circumstances \cite{PX4-external-modes-limitations}.

ROSflight avoids this complexity by allowing all control loops to be closed on the Linux-based companion computer, thus improving flexibility. 
Use cases might include sending direct actuator commands for individual motor control during transition of an eVTOL, or for sending actuator commands from the output of a neural network controller \cite{reinforcement_uav_control,reinforcement_crazyflie}.
Additionally, operating ROSflight in pass-through mode enables users to filter the inputs to the mixer.
This could be used to simulate hardware failures \cite{physical_failure}, cyber attacks \cite{cyber_attack}, or other physical failures simply by filtering the companion computer's output before sending to the mixer.

\subsubsection{Primary and Secondary Mixers}
The addition of a custom mixer can raise issues for a safety pilot.
For example, a quadrotor with the identity matrix as the mixing matrix would allow the companion computer to send actuator commands directly to each of the four motors.
This scenario, however, would make it impossible for a safety pilot to fly the quadrotor, because the rate and angle controllers used by the RC pilot output forces and torques, not direct actuator commands, as shown in Figure \ref{fig:mixer-diagram}.

To address this safety concern, ROSflight supports loading two separate mixers: the primary mixer and the secondary mixer.
The primary mixer is used by the RC safety pilot and thus should always be defined such that the RC pilot can control the vehicle, while the secondary mixer is used by the companion computer.
When the companion computer has control over the vehicle, ROSflight uses the secondary mixer; otherwise, it uses the primary mixer.

\begin{figure}
    \centering
    \includegraphics[width=1.0\columnwidth]{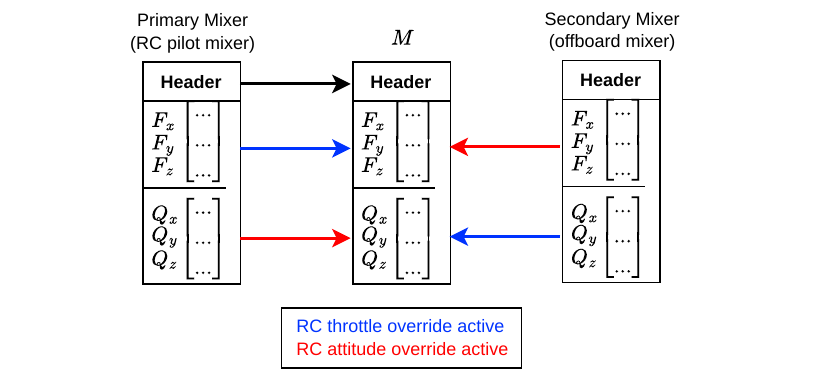}
    \caption{Diagram of how the ROSflight mixer, $M$, is constructed from the primary and secondary mixers based on which RC overrides are active. Note that the header information (PWM rate and output type) for $M$ is always constructed from the primary mixer's header information.}
    \label{fig:pri_vs_sec_mixer}
\end{figure}

When conducting flight tests with a companion computer, it can be helpful to isolate certain channels of the offboard command to incrementally test companion computer control.
To accommodate this, the RC pilot in ROSflight has two types of control over the vehicle: attitude override and throttle override.
When RC attitude override is enabled, the RC pilot has control over the attitude ($Q$) portions of the command vector $u$.
When RC throttle override is enabled, the RC pilot has control over the throttle ($F$) portions of the command vector.
When either override is enabled, the mixer that transforms the control input vector $u = [F_x, F_y, F_z, Q_x, Q_y, Q_z, 0_{1\times4}]^T$ to motor outputs $\tau$ is a mixture of the primary and secondary mixers, as shown in Figure (\ref{fig:pri_vs_sec_mixer}).
This allows the RC pilot to control either the throttle or the attitude while the companion computer controls the other.
Note that this requires that the mixing matrices and the control inputs are designed so that this split may occur.
In other words, if the mixer does not have proper meaning as a combination of the primary and secondary mixers, then attempting to use attitude and throttle override independently of each other may cause undesirable behavior.
If that is the case, setting the attitude and throttle override RC channels to the same value (thus activating both or neither) would result in either the entire primary mixer or the entire secondary mixer being used.

Both the primary mixer and the secondary mixer should be selected by setting the appropriate parameter through the ROSflight parameter interface. 
If the secondary mixer is left unspecified, it will default to the same value as the primary mixer.
This is useful for users who do not need a distinction between the primary and secondary mixers.

\subsubsection{Empirical Motor Parameters}
As described in \cite{rosflight_arxiv}, using the motor and propeller parameters can increase the fidelity of the mixing matrix.
In ROSflight, the \texttt{USE\_MOTOR\_PARAM} parameter tells the firmware whether or not the mixer directly outputs motor setpoints, $\delta_{t,i}$, or desired angular speeds, $\Omega_i$.
If the mixer outputs desired angular speeds, then ROSflight converts these values to desired motor setpoints before they are sent to the electronic speed controllers (ESC).

\section{Hardware Integration}\label{hardware-updates}
\subsection{Hardware Support}

% Part Inventory
% Config 1: 
%    - Pixracer Pro M10064D. https://store.3dr.com/pixracer-pro/
%    - Power board: MRO M10121A High Current Power Module
%    - GPS/mag: MRO M10034C GPS board, has Ublox NEO-M9N GPS chip with IST8308 3-axis mag
%    - Pitot board: MRO I2C Airspeed Sensor MS4525DO
%    - Jetson Orin NX: Jetson Orin NX 8GB Developer Kit, OR Raspberry Pi 5 8GB
% Config 2:
%    - Varmint 10/11X
%    - Jetson Orin NX 16 GB
%    - M.2 SSD: Western Digital IX SN530 NVMe, 512GB
%    - GPS antenna: 1575 MHz GPS active antenna
% Common:
%    - RC receiver: TBS Crossfire 8Ch Diversity Rx 
%    - RC transmitter: TBS Crossfire TX Lite
%    - Doodle labs receiver: Doodle Labs 5GHz Embedded Mesh Rider, RM-5800-2J-XM
%    - Doodle labs transmitter: Doodle Labs 5GHz Smart Radio, RM-5800-2J-XE
% Notes:
% These RC transmitters and receivers could be replaced by any generic RC pair that outputs SBUS on the vehicle side.
% The WiFi transmitter/radio could be replaced by any radio. We have used Ubiquiti Bullets and Rockets before, but for ROSflight, we have mainly stuck with the Doodle Labs.

\begin{figure}
    \centering
    \includegraphics[width=\columnwidth]{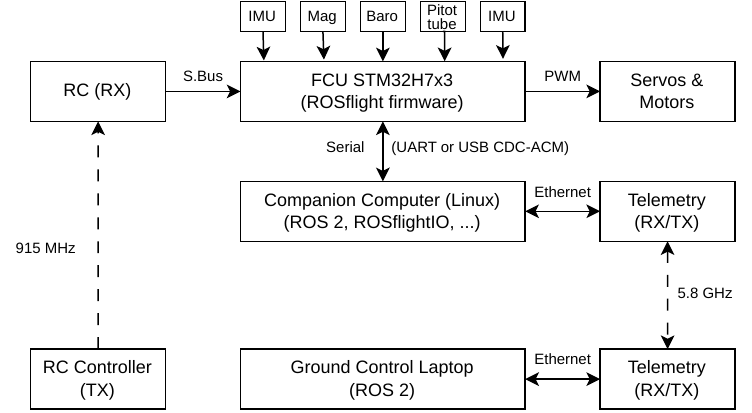}
    \caption{Hardware elements}
    \label{fig:hardware-comp-elements} 
\end{figure}

ROSflight supports FCU hardware configurations having at a minimum a six-axis inertial measurement unit (IMU with three-axis accelerometer + three-axis gyroscope) for local state estimation and control, and a radio control (RC) receiver for some control inputs.
Here, we employ FCUs that additionally include three-axis magnetometer, barometer, differential pressure, and GPS sensors; and a telemetry data link.
Measurement of this additional sensor data via the FCU ensures consistent and accurate relative timestamps for use by the companion computer.
Figure \ref{fig:hardware-comp-elements} illustrates the interconnection of components, including the FCU, companion computer, telemetry, and ground control elements. 

Two hardware configurations are currently supported and demonstrated (Table \ref{tab:hardware-comp-list}). 
Configuration 1 is comprised of commercial off the shelf (COTS) components from 3DR and Nvidia, and is therefore more accessible.
Configuration 2, provided by AeroVironment, Inc. (AV) benefits from a more tightly integrated hardware package, as well as higher performance IMU and differential pressure sensors.
Both configurations share compatible STM32H7 microcontrollers allowing significant software commonality.
RC and telemetry data links used in hardware experiments are listed in Table \ref{tab:hardware-comp-list-common-telemetry}.
Note that other RC and telemetry data links are also supported.

To support different hardware (e.g., GPS, magnetometer, other sensor boards, or a different FCU) and to improve adaptability to different hardware configurations, 
ROSflight is designed so that the core firmware functionality is abstracted from the hardware implementation, called the hardware abstraction layer (HAL).
The HAL for a given hardware configuration is created by inheriting from a base class that defines all the functions specifically related to hardware required by the ROSflight firmware.
For example, to support a different GPS receiver, a user would create a driver to interface with the new receiver and then interface with ROSflight using the associated class member functions defined by the HAL.
More generally, the entire board can be replaced as long as it employs the board HAL. 

\begin{table}
    \begin{subtable}[h]{0.45\textwidth}
        \centering
        \label{tab:hardware-comp-list-config1}        
        \caption{FCU configuration 1}
        \begin{tabular}{>{\hangindent=.1in}p{0.3\columnwidth}|p{0.55\columnwidth}}
            \toprule
              Component & Function \\
             \midrule
             3DR Pixracer Pro & FCU with IMU and barometer\\
             3DR M10034C & GPS and magnetometer \\
             3DR M10121A & Power board \\
             3DR MS4525DO & Differential pressure \\
             Nvidia Jetson Orin Nano 8GB developer kit OR Raspberry Pi 5 8GB & Companion computer \\
            \bottomrule
        \end{tabular}
        \vspace{3mm} 
    \end{subtable}
    \begin{subtable}[h]{0.45\textwidth}
        \centering
        \caption{FCU configuration 2}
        \label{tab:hardware-comp-list-config2}
        \begin{tabular}{p{0.3\columnwidth}|>{\hangindent=0.1in}p{0.55\columnwidth}}
            \toprule
              Component & Function \\
             \midrule
             AV Varmint & FCU with integrated IMU, barometer, magnetometer, differential pressure, GPS, power, and Nvidia Jetson Orin NX 16GB companion computer\\
             1575A-L & GPS L1 active antenna\\ 
            \bottomrule
        \end{tabular}
        \vspace{3mm} 
    \end{subtable}
    \caption{Controller configurations}
    \label{tab:hardware-comp-list}
\end{table}

\begin{table}
    \centering
    \begin{tabular}{>{\hangindent=0.1in}p{0.6\columnwidth}|p{0.25\columnwidth}}
        \toprule
          Component & Function \\
         \midrule
         TBS Crossfire 8Ch Diversity Rx & RC receiver \\
         TBS Crossfire TX & RC transmitter \\
         Doodle Labs 5GHz Embedded Mesh Rider, RM-5800-2J-XM & Air telemetry \\
         Doodle Labs 5GHz Smart Radio, RM-5800-2J-XE & Ground telemetry \\
        \bottomrule
    \end{tabular}
    \caption{Remote Control and Telemetry}
    \label{tab:hardware-comp-list-common-telemetry}  
    \vspace{-15pt}
\end{table}

\subsection{Software-in-the-Loop Simulation} \label{sec:simulation}
A simulator is an essential component in robotics research that allows for rapid development before deploying on hardware.
In many cases, the transition from simulation to hardware experiments can require extensive modifications or adjustments to the software used for the experiments due to hardware or other real-world constraints.
ROSflight seeks to minimize the effort to transition from simulation to hardware environments by mirroring the hardware environment as closely as possible in simulation.
This enables the same software that runs in simulation (i.e., the ROSflight firmware and an associated autonomy stack) to also control the vehicle in hardware, with no changes.

Many simulation environments have been developed to aid UAV research, including general-purpose simulators and application-specific simulators \cite{simulator-survey}.
Each simulator has different attributes that make it more or less suited to a given research problem.
For example, research that uses computer vision as an essential component of the autonomy stack likely needs a photorealistic simulator, while research focusing on multi-agent coordination and localization may only need a headless environment to do Monte-Carlo experiments.

\begin{figure}
    \centering
    \captionsetup[subfigure]{skip=4pt}
    \subcaptionbox{Multirotor flying a waypoint mission in the RViz simulation environment.}
        {\includegraphics[width=\columnwidth]{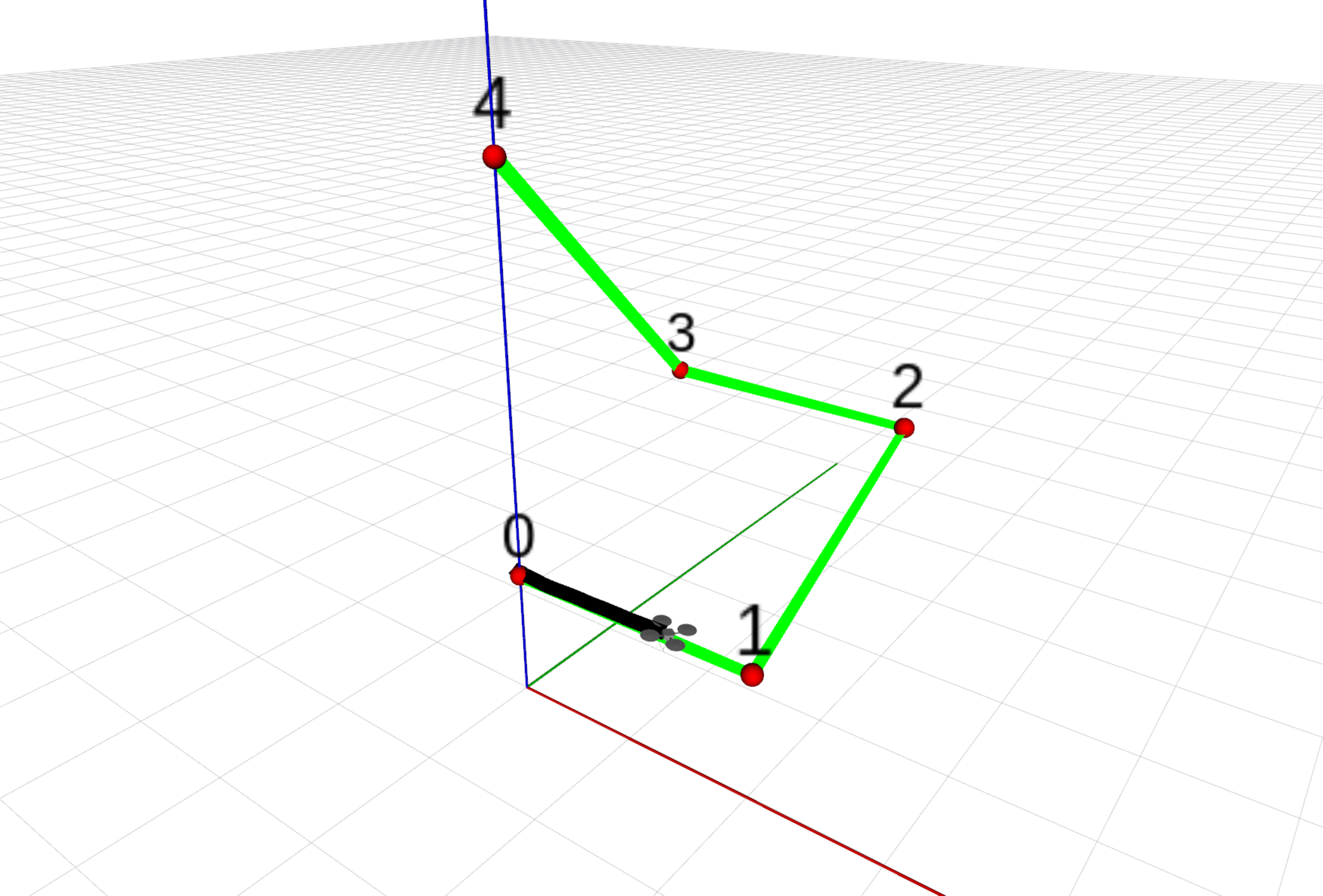}}
    \vspace{10pt}
        
    \captionsetup[subfigure]{skip=4pt}
    \subcaptionbox{Fixed-wing aircraft flying a waypoint mission in the Gazebo simulation environment.}
        {\includegraphics[width=\columnwidth]{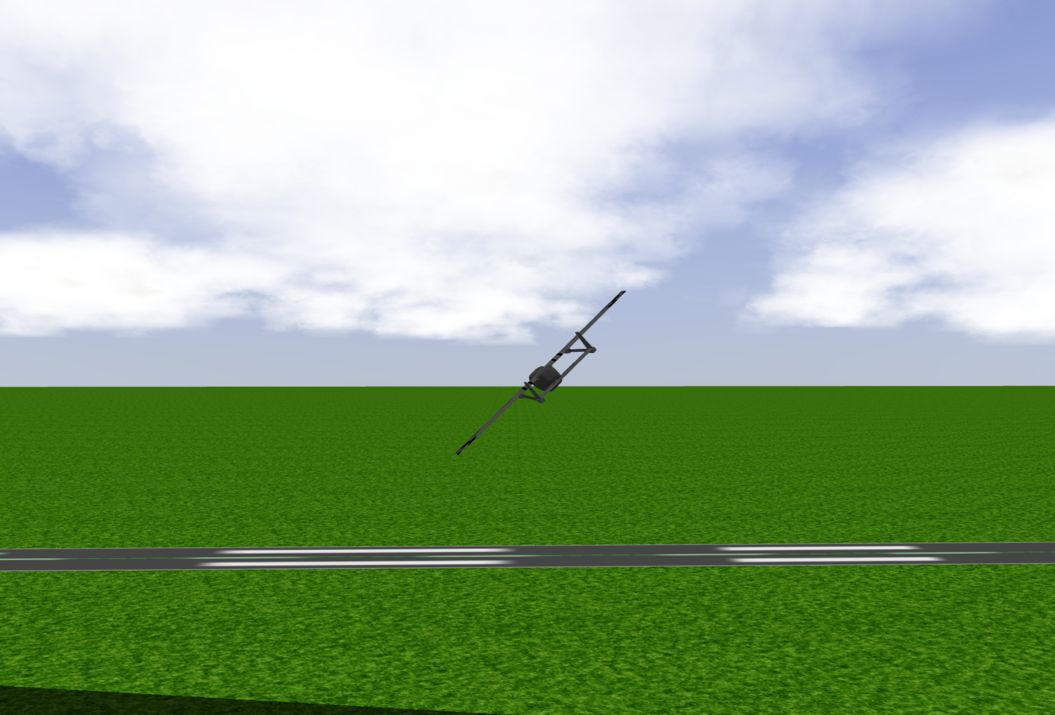}}
    \vspace{10pt}
        
    \captionsetup[subfigure]{skip=4pt}
    \subcaptionbox{Multirotor flying in the photorealistic Unreal Engine 5-based HoloOcean simulation environment. Camera images can be easily simulated.}
        {\includegraphics[width=\columnwidth]{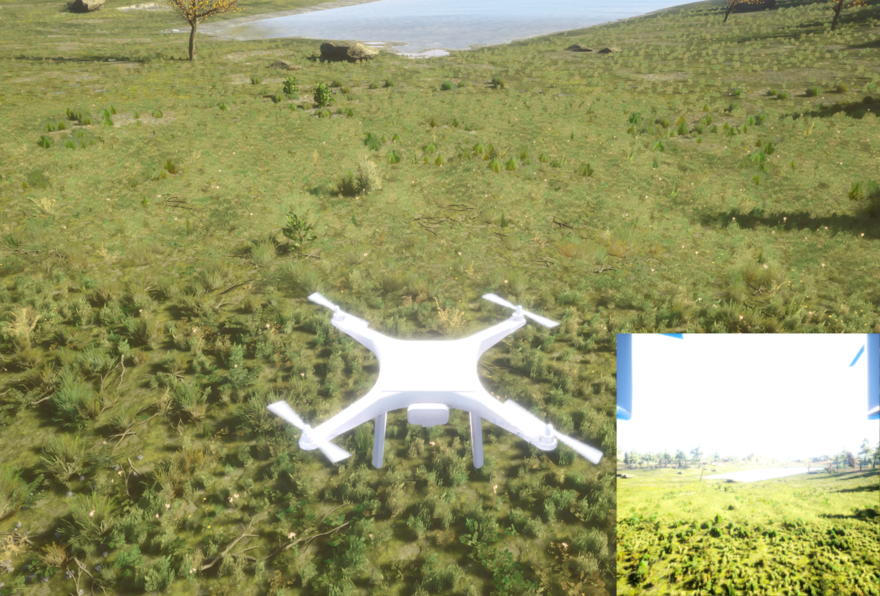}}
    
    \caption{Simulation environments supported natively by ROSflight, from photorealistic to lightweight environments that can be run headlessly.}
    \label{fig:simulation-envs}
\end{figure}

ROSflight natively supports three simulators out of the box: HoloOcean \cite{holoocean}, a photorealistic visualizer built on Unreal Engine 5, Gazebo \cite{gazebo}, a popular open-source simulator for robotics, and a custom lightweight simulator that uses ROS 2 RViz for visualization.
Figure (\ref{fig:simulation-envs}) shows each of these three supported environments.

\begin{figure}
    \centering
    \includegraphics[width=0.9\linewidth]{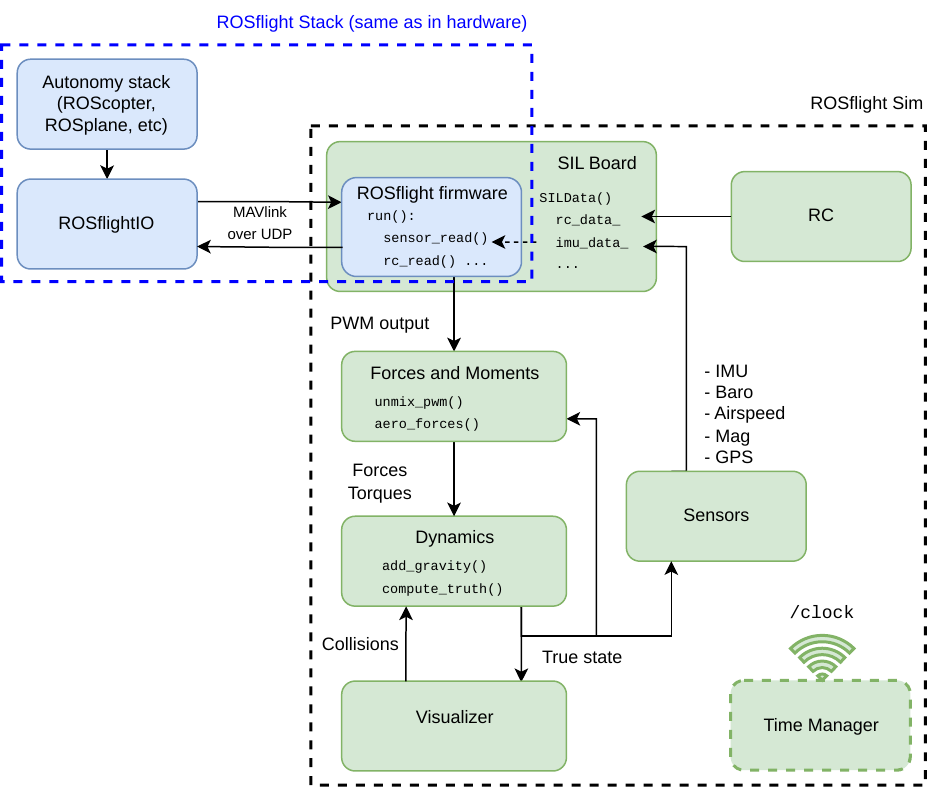}
    \caption{Diagram of the modules of the ROSflight simulation. Each module has a single responsibility, enabling easier modification by reducing dependencies between modules.}
    \label{fig:rosflight_sim_arch}
\end{figure}

Since different projects often require simulation environments with different features, the ROSflight simulation package is flexible and modular so that users can support their own simulator or include custom functionality.
To do this, the ROSflight simulation is split into a series of modules which are implemented as separate ROS 2 nodes and are described in Table \ref{table:sim_arch}.
Figure (\ref{fig:rosflight_sim_arch}) shows how each module interacts and what information is passed between modules.
This modularity enables users to define which modules are loaded at runtime.
Since different visualizers come with different capabilities, this allows visualizers to be exchanged with minimal effort.
For example, Gazebo performs dynamic propagation as part of the visualization engine, while the lightweight RViz visualizer only displays the 3D motion of the aircraft.
Thus, when using Gazebo with ROSflight, the dynamics module is not launched while the rest of the ROSflight simulation modules remain unchanged.

\begin{table}
    \centering
    \begin{tabular}{p{0.3\columnwidth}|>{\hangindent=0.1in}p{0.55\columnwidth}}
        \toprule
        \textbf{Module Name} & \textbf{Description} \\
        \midrule
        Time Manager & Manages simulation time \\
        SIL Board & Instantiation of firmware and simulated FCU board \\
        RC & Simulates RC safety pilot connection \\
        Sensors & Creates simulated sensor measurements \\
        Forces and Moments & Computes aerodynamic forces and moments \\
        Dynamics & Handles dynamic integration, and manages truth and environment state \\
        Visualizer & Plots the vehicle in the simulated environment \\
        \bottomrule
    \end{tabular}
    \caption{Simulation Module Descriptions}
    \label{table:sim_arch}
    \vspace{-15pt}
 \end{table}
 
The separation of responsibilities into modules also enables easy customization of each module.
For example, eVTOL AAM aircraft are often quad-plane or tilt-rotor type vehicles and, as such, have nonstandard aerodynamic and propulsion models.
The responsibility of the forces and moments module in ROSflight is to compute the aerodynamic and propulsive forces and moments from the input actuator commands.
Users wishing to use a different aerodynamic model need only to replace the ROSflight forces and moments module, while the rest of the simulation environment remains unchanged.
Sensors can also be added and customized with minimal effort.
The sensors node is responsible for creating simulated sensor measurements based off of the true state of the vehicle, and includes an IMU, GNSS, barometer, magnetometer, sonar, and differential pressure sensor by default.
To add a new sensor, one simply needs to create a ROS 2 publisher that generates the desired sensor data at a given rate.

Each module in the ROSflight simulation package is defined by an interface class. 
This interface class defines the ROS 2 interfaces (e.g., publishers, subscribers, etc.) that a module has, and defines virtual functions that must be implemented by derived classes.
As long as a node inherits from the interface class and implements the required functionality, that node will interact with the rest of the ROSflight simulation environment.
Thus, other simulators can easily be supported by writing a ROS 2 wrapper around the simulator's API---as long as the new wrapper has the same interfaces as the interface class, the new simulator will interact properly with the rest of the ROSflight simulation.

\section{Hardware Demonstrations}\label{hardware-demos}
Hardware experiments were performed to validate the improvements made to ROSflight 2.0.
In the first, we measure the approximate serial delays from sending offboard commands to ROSflight from the companion computer.
In the second, we demonstrate ROSflight in pass-through mode controlling a multirotor at 400 Hz.

\subsection{Serial Delay}
To measure the approximate serial delay, we modified ROSflight firmware to send all received offboard commands back to the companion computer.
The companion computer first packed an offboard command message, marked the timestamp, and sent it to ROSflight firmware.
The firmware parsed the message as part of its normal routine and immediately sent it back.
Thus, each offboard command was sent twice across the serial connection.
The companion computer received the message and computed the round-trip time (RTT).
Note that the ROSflight firmware continued to run all of its normal operations during this experiment, including collecting, packing, and sending sensor data, responding to heartbeat requests, filling parameter requests, etc.
This means that some variability and delay is included in the RTT measurement, as well as the time required for the normal operation of ROSflight.
Data was collected for 45 seconds, and the average, maximum, and minimum RTT values were recorded.
Table \ref{tab:config2-bench-serial-slow} shows RTT statistics for when the companion computer sends offboard commands at 400 Hz.
A Raspberry Pi 5 was used as the companion computer for all configuration 1 tests.
\begin{table}
    \centering
    \begin{tabular}{c|ccc}
        \toprule
         Config. &  Ave (ms) & Max (ms) & Min (ms) \\ 
         \midrule
         1 & 1.712 & 84.103 & 0.146 \\ % 18476 samples
         2 & 0.416 & 2.334 & 0.174 \\ % 19394 samples
         \bottomrule
    \end{tabular}
    \caption{Bench test of RTT and publishing rate for both hardware configurations. The companion computer published ROS 2 offboard commands at 400 Hz.}
    \label{tab:config2-bench-serial-slow}
    \vspace{-15pt}
\end{table}

Table \ref{tab:config2-bench-serial-fast} shows RTT statistics for when the companion computer sent offboard commands as fast as possible.
To do this, ROS 2 commands were published at a configurable rate, which was increased until the companion computer was unable publish ROS 2 messages fast enough.
The maximum rate for both configurations was above 1100 Hz.
The size of the offboard command message is 24 bytes, meaning that the serial connection was able to send $>52800$ bytes per second of offboard command information under normal operation of ROSflight.
\begin{table}
    \begin{tabular}{c|cccc}
        \toprule
         Config. &  Ave (ms) & Max (ms) & Min (ms) & \shortstack{Average received \\ command rate (Hz)}\\ 
         \midrule
         1 & 1.024 & 56.921 & 0.154 & 1140 \\ % 52440 samples
         2 & 0.385 & 3.095 & 0.166 & 1360 \\ % 62504 samples
         \bottomrule
    \end{tabular}
    \caption{Bench test of RTT and publishing rate for both hardware configurations. The companion computer published commands over the ROS 2 network as fast as possible.}
    \label{tab:config2-bench-serial-fast}
    \vspace{-15pt}
\end{table}

\begin{figure}
    \centering
    \includegraphics[width=0.9\linewidth]{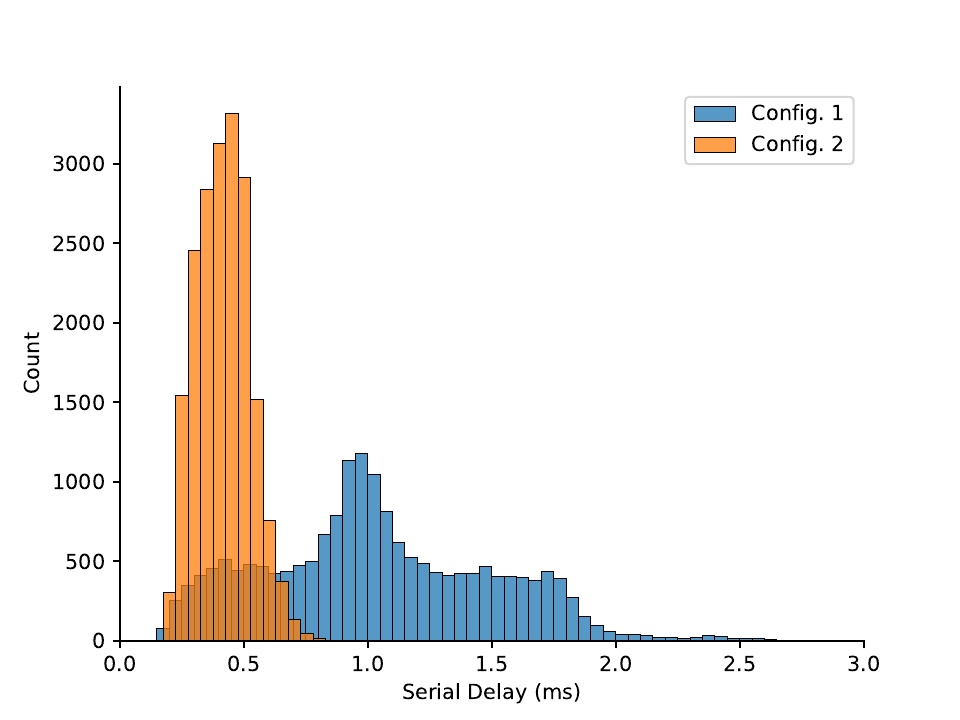}
    \caption{Recorded RTT for offboard commands published at 400 Hz for both hardware configurations.}
    \label{fig:rtt_comparison}
    \vspace{-15pt}
\end{figure}

Figure \ref{fig:rtt_comparison} shows a histogram of the recorded RTT values for the data presented in Table \ref{tab:config2-bench-serial-slow}, with commands being sent at 400 Hz.
This shows that the maximum value of 56.921 ms reported in Table \ref{tab:config2-bench-serial-slow} for configuration 1 is an outlier.
Configuration 2 demonstrated significantly superior RTT values over configuration 1.
Note that configuration 2 has an integrated design, but configuration 1 relies on a physical USB connection for the serial connection.

\subsection{ROSflight in Pass-through}
A flight test of ROSflight flying a multirotor in pass-through mode was conducted.
For this experiment, configuration 2 was used on a Holybro x650 quadrotor frame.
A custom mixer was computed using the proposed methods.
A ROS 2 node containing a PID-based angle controller was implemented on the companion computer, which output forces and torques.
These force and torque offboard commands were published over the ROS 2 network to the ROSflightIO node at 400 Hz, which forwarded them to ROSflight over the serial connection.
Since ROSflight was operating in pass-through mode, these offboard commands did not pass through either of ROSflight's controllers, but instead progressed directly to the mixer.
This operation is equivalent to sending direct motor commands from the companion computer to ROSflight over the serial connection.

\begin{figure}
    \centering
    \includegraphics[width=0.9\linewidth]{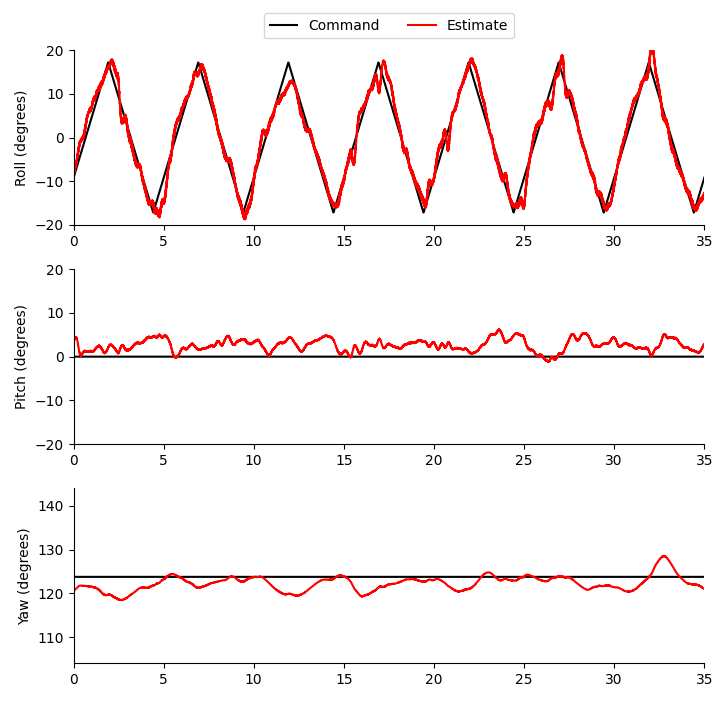}
    \caption{Multirotor response in hardware to roll triangle-wave setpoints with the onboard computer sending forces and torques to the ROSflight mixer in pass-through mode, under mild wind conditions. Configuration 2 was used for this experiment.}
    \label{fig:roll-cmd-experiment}
    \vspace{-15pt}
\end{figure}

The system's response to a triangle-wave roll command is shown in Figure \ref{fig:roll-cmd-experiment}, demonstrating that the companion computer was able to control the multirotor over a serial connection at 400 Hz with all control loops on the companion computer.
We emphasize that the offboard commands were sent directly to the ROSflight mixer, bypassing all control loops on the FCU. 

\section{Summary}\label{summary}
In this work we present significant advancements to ROSflight that improve the usability and modularity of the system.
These advancements include the transition from ROS 1 to ROS 2, customizable actuator mixing, new supported hardware, and a more modular simulation environment.
These improvements enable ROSflight to accelerate research in areas like advanced air mobility and UAVs.
We demonstrated these new improvements in hardware, using ROSflight to control a multirotor at 400 Hz in pass-through mode, meaning all control loops were closed on the Linux-based companion computer.
Approximate round-trip time delays in the serial connection were presented for both of the supported hardware configurations.
Because ROSflight moves most of the higher-level autonomy functions to the companion computer, the ability to operate ROSflight in pass-through mode enables faster development and access to greater compute resources.
ROSflight is documented in greater detail on the project website, \href{rosflight.org}{rosflight.org}, where users can contribute or ask questions.

\appendix
This appendix contains the derivation of the general-form and predefined mixer equations for standard multirotor vehicles used in ROSflight.
This derivation assumes that all outputs are motor outputs, not servo outputs.
The fixed-wing mixer derivation is not included here, but the mixer equations for a general aircraft can be derived by following a similar procedure. 

\subsection{General Form}
As defined in \cite{uavbook}, the thrust and torque generated by a motor and propeller in vector form are
\begin{equation}\label{eq:A:thrust-mixer}
    \Vec{F}_{i} = C_T \frac{\rho D^4}{4 \pi^2} \Omega_i^2 \hat{e}_i
\end{equation}
\begin{equation}\label{eq:A:torque-mixer}
\begin{split}
\Vec{Q}_i & = \vec{r} \times \vec{F}_i + C_Q \frac{\rho D^5}{4 \pi^2} \Omega_i^2 d_i \hat{e}_i \\
          & = C_T \frac{\rho D^4}{4 \pi^2} \Omega_i^2 (\vec{r} \times \hat{e}_i)  + C_Q \frac{\rho D^5}{4 \pi^2} \Omega_i^2 d_i \hat{e}_i,
\end{split}
\end{equation}
where $\hat{e}_i$ is the unit vector pointed in the direction of the axis of rotation of the motor, $C_T$ and $C_Q$ are the thrust and torque coefficients associated with the propeller (assumed to be constant), $\rho$ is the air density, $D$ is the propeller diameter, $\vec{r}$ is the vector pointing from the aircraft center of mass to the motor's axis of rotation, $d_i \in \{-1,1\}$ encodes the direction of rotation of a propeller, and $\Omega_i$ is the angular velocity of the propeller and motor.

We wish to rewrite Equations (\ref{eq:A:thrust-mixer}) and (\ref{eq:A:torque-mixer}) using Equation (\ref{eq:mixer}) to derive the mixing matrix $M$.
Then, the desired forces and torques relate to the desired angular velocities squared by the mixing matrix $M$ according to
\[
\begin{bmatrix}
\Vec{F}_d \\ \Vec{Q}_d \\
\end{bmatrix}
=
M
\begin{bmatrix}
    \Omega_1^2 \\ 
    \Omega_2^2 \\
    \vdots \\
    \Omega_n^2 \\
\end{bmatrix}
=
\sum_{i=1}^n M_{i} \Omega_i^2
=
\sum_{i=1}^n
\begin{bmatrix}
    \Vec{F}_i \\
    \Vec{Q}_i \\
\end{bmatrix}
\]
where $\Vec{F}_d$ and $\Vec{Q}_d$ are the desired thrust and torque, $M_i$ is the $i$th column of $M$, and $\Vec{F}_i$ and $\Vec{Q}_i$ are the thrust and torque produced by the $i$th motor.

The $i^{\text{th}}$ column of $M$ is then given by
\[
\begin{bmatrix}
\Vec{F}_i \\ \Vec{Q}_i \\
\end{bmatrix}
=
M_i \Omega_i^2
=
\underbrace{
C_T \frac{\rho D^4}{4 \pi^2}
\begin{bmatrix}
    \hat{e}_i \\
    \vec{r} \times \hat{e}_i  + \frac{C_Q D d_i}{C_T} \hat{e}_i \\
\end{bmatrix}
}_{M_i}
\Omega_i^2.
\]
This is a general form, so it is valid for any motor configuration, and the desired angular speeds can be computed by left-multiplying both sides by $M^\dagger$, where $(\cdot)^\dagger$ is the Moore-Penrose pseudoinverse.

If we know the motor parameters, we can compute the desired input voltage based on the desired angular velocities.
From \cite{uavbook}, we have
\begin{equation}
    Q_p = \frac{\rho D^5}{4\pi^2}\Omega_i^2 C_Q
\end{equation}
\begin{equation}
    Q_m = K_Q \left[ \frac{1}{R}(V_\text{in} - K_V\Omega_i - i_0 \right],
\end{equation}
where $Q_p$ is the torque generated by the propeller due to air resistance, $Q_m$ is the torque generated by the motor, $K_Q$ is the motor torque constant, $R$ is the resistance of the motor, $K_V$ is the back-EMF voltage constant, and $i_0$ is the no-load current.

Setting $Q_m = Q_p$ (as is true under steady-state conditions) yields
\begin{equation}\label{eq:A:vin}
V_{\text{in}} = \frac{R C_Q}{K_Q} \frac{\rho D^5 }{4 \pi^2} \Omega_i^2 + i_0 R + K_V \Omega_i. 
\end{equation}
The throttle setting, $\delta_{t,i} \in [0,1]$ can then be calculated using $V_{\text{in}} = \delta_{t,i} V_{\text{max}}$.
This throttle setting is the PWM setting that the mixer writes to the motors, assuming that output voltage is scaled linearly to the PWM duty-cycle setting.

\subsection{Simplifications}
The general form of the mixing matrix described above requires knowledge of the motor parameters of the system.
Some simplifications can be made to generate mixing matrices that do not depend on the motor and propeller parameters.
This results in a more intuitive and simple mixing matrix, but requires different controller gains on the firmware controller to achieve the same performance.

\noindent
\textbf{Assumption 1:} All motors and propellers have the same thrust and torque coefficients.

\noindent
\textbf{Assumption 2:} $\hat{e}_i = -\hat{k} = \begin{bmatrix}
    0 & 0 & -1
\end{bmatrix}^T$ in body-fixed north-east-down (NED) frame, meaning that the motors are all oriented in the $-z$ direction (i.e., thrusting straight up).

Then, 
\[
\vec{r} \times \hat{e}_i
=
\begin{bmatrix}
    r_y \hat{e}_{i,z} - r_z \hat{e}_{i,y} \\
    r_z \hat{e}_{i,x} - r_x \hat{e}_{i,z} \\
    r_x \hat{e}_{i,y} - r_y \hat{e}_{i,x} \\
\end{bmatrix}
=
\begin{bmatrix}
    -r_y \\
    r_x \\
    0 \\ 
\end{bmatrix}
=
\begin{bmatrix}
    -||\vec{r}||\sin\theta \\
    ||\vec{r}||\cos\theta \\
    0 \\ 
\end{bmatrix}
\]
and 
\begin{equation} \label{eq:A:assump2}
\begin{bmatrix}
    \Vec{T}_i \\ \Vec{Q}_i \\
\end{bmatrix}
=
C_T \frac{\rho D^4}{4 \pi^2}
\begin{bmatrix}
    0 \\ 
    0 \\
    -1 \\ 
    -||\vec{r}||\sin\theta \\
    ||\vec{r}||\cos\theta \\
    \frac{C_Q D}{C_T} d_i \\
\end{bmatrix}
\Omega_i^2,
\end{equation}
where $\theta$ is the angle from the positive body-fixed x-axis, as in \cite{uavbook}.

\noindent
\textbf{Assumption 3:} Each desired input value is computed independently (i.e., separate gains in a controller) than the other input values, and $||\vec{r}||$ is constant for all motors.

Then, the constant terms in (\ref{eq:A:assump2}) can be factored out and subsumed into the controller gains (and thus the desired inputs), leading to
\[
\begin{bmatrix}
    \Vec{T}_i \\ \Vec{Q}_i \\
\end{bmatrix}
=
\begin{bmatrix}
    0 &
    0 &
    -1 &
    -\sin\theta &
    \cos\theta &
    d_i
\end{bmatrix}^T
\Omega_i^2.
\]

\noindent
\textbf{Assumption 4:} $V_{\text{in}} \approx \frac{R \rho D^5 C_Q}{4 \pi^2 K_Q} \Omega_i^2$, meaning that the squared term in (\ref{eq:A:vin}) dominates.

The constants in \ref{eq:A:vin} can be subsumed as above.
Furthermore, since $V_{\text{in}} = \delta_{t,i} V_{\text{max}}$ we have 
\begin{equation}\label{eq:A:simplified_mixer}
\begin{bmatrix}
    \Vec{T}_i \\ \Vec{Q}_i \\
\end{bmatrix}
=
\begin{bmatrix}
    0 &
    0 &
    1 &
    -\sin\theta &
    \cos\theta &
    d_i
\end{bmatrix}^T
\delta_{t,i}.
\end{equation}

Equation (\ref{eq:A:simplified_mixer}) shows the $i$th column of a simplified $M$, where the rows correspond to geometric properties associated with the given airframe.
The assumptions made in the derivation of this equation limit the accuracy of this mixer model, but allow for an easy and intuitive understanding of the mixing matrix.
The hard-coded mixing matrices (for multirotors) in ROSflight follow this form, allowing users to easily determine the mixing matrix and its limitations without knowledge of the motor parameters of the system.

\bibliographystyle{IEEEtran}
\bibliography{bibi}

\end{document}